\documentclass{article}
\usepackage{graphicx} 
\usepackage[english]{babel}
\usepackage{amsmath}
\usepackage{amsfonts}
\usepackage{amssymb}
\usepackage{float}
\usepackage{hyperref}
\usepackage{booktabs}
\usepackage{multirow}

\begin{document}

\title{SDPM: Survival Diffusion Probabilistic Model for Continuous-Time Survival Analysis}
\author{Stanislav R. Kirpichenko, Andrei V. Konstantinov, Lev V. Utkin \\ 
\small{Peter the Great St.Petersburg Polytechnic University} \\ 
\small{kirpich\_sr@spbstu.ru, konstantinov\_av@spbstu.ru, utkin\_lv@spbstu.ru}}
\date{}
\maketitle

\begin{abstract}
Survival analysis aims to estimate the distribution of time-to-event outcomes from data with censored observations. Many existing methods either impose structural assumptions on the hazard function or discretize the time axis, which may limit flexibility and introduce approximation errors.

In this paper, we propose the Survival Diffusion Probabilistic Model (SDPM), a generative approach to continuous-time survival analysis. SDPM models the conditional distribution of the observed survival outcome, represented by the pair of observed time and censoring indicator, $\mathbb{P}(T,\delta \mid \mathbf{x})$, using a denoising diffusion probabilistic model. Under the standard assumption of conditionally independent censoring, conditional samples generated by the model can be transformed into survival function estimates using the Kaplan-Meier estimator. This formulation avoids explicit parametric assumptions on the event-time distribution and does not require a fixed discretization of the output time space.

The proposed model operates in a transformed target space, using standardized log-times and a continuous Gaussian-mixture representation of the censoring indicator, which improves numerical stability and sample validity. We evaluate SDPM on ten real-world survival datasets and compare it with five strong baselines, including tree-based, boosting-based, and neural survival models. The results show that SDPM achieves competitive predictive performance across Harrell's C-index, integrated time-dependent AUC, and integrated Brier score. In particular, the method obtains the best average rank among the compared approaches and demonstrates especially strong performance in terms of IBS, indicating accurate survival function estimation.

Additional experiments analyze the main properties of the proposed generative formulation. We show that increasing the number of generated samples improves the stability and fidelity of the Kaplan-Meier reconstruction, while the model remains reasonably robust to moderate changes in the number of reverse diffusion steps. A qualitative and quantitative study on synthetic Cox-Weibull data demonstrates that SDPM can recover the shape of an underlying continuous survival distribution more accurately than a strong nonparametric baseline when sufficiently many samples are generated. Finally, an ablation study confirms the importance of the proposed target-space transformations, which improve event-rate calibration, reduce invalid generated times, and provide consistent gains in predictive discrimination. Codes implementing the proposed model are publicly available


\end{abstract}

\section{Introduction}

Survival analysis is an important tool in a wide range of applied domains, including medicine, reliability engineering, and financial modeling. The central task is to estimate the distribution of time to an event from observed data containing both fully observed and censored outcomes.

Most modern survival analysis methods can be roughly divided into three groups. Classical statistical approaches, such as the Cox proportional hazards model, rely on assumptions about the structure of the hazard function. Nonparametric methods, including random survival forests, provide greater flexibility but do not yield an explicit probabilistic model of event-time distributions. Modern deep learning methods, such as DeepSurv and DeepHit, can capture complex nonlinear relationships between features, but they either retain some of the limitations of classical models or require discretization of the time axis.

Time discretization, which is widely used in neural survival models, introduces several limitations: the choice of grid affects approximation accuracy, the dimensionality of the problem increases, and a trade-off arises between precision and computational complexity. In addition, many existing methods do not directly model the joint distribution of event time and censoring indicator, which restricts their expressive power.

In this work, we propose a new generative approach to survival analysis based on modeling the conditional joint distribution of the observed time and censoring indicator, $\mathbb{P}(T, \delta \mid \mathbf{x})$. To this end, we employ a diffusion probabilistic model that generates samples of pairs $(T, \delta)$ consistent with the conditional distribution for a given feature vector.

The key idea is that survival function estimation can be reduced to sampling from a conditional distribution followed by nonparametric Kaplan-Meier estimation. This approach avoids explicit parameterization of the event-time distribution and naturally operates in continuous time without requiring a fixed discretization of the output space.

The proposed approach has the following advantages:
\begin{itemize}
    \item modeling the joint distribution of event time and censoring indicator;
    \item operation in continuous time without requiring a fixed discretization of the time axis;
    \item controllable accuracy of the survival estimate via the number of generated realizations;
    \item the ability to recover the shape of the underlying survival distribution from generated samples;
    \item combining generative modeling with classical nonparametric methods for survival analysis.
\end{itemize}

In this paper, we introduce the Survival Diffusion Probabilistic Model (SDPM), a generative approach to survival analysis that models the joint distribution of event time and censoring indicator in continuous time. We describe its architecture and training procedure, and demonstrate through extensive experiments that the proposed method provides competitive predictive performance, controllable estimation accuracy, and the ability to recover the underlying continuous survival distribution.

Codes implementing the proposed model are publicly available
 at: \url{https://github.com/NTAILab/survival_diffusion}.

\section{Related Work}

\textbf{Survival Analysis in Machine Learning}
Survival analysis has become increasingly important in machine learning due to its applications across medicine, reliability engineering, financial modeling, and many other domains. Comprehensive reviews of survival models in machine learning have been provided in recent surveys \cite{Wang-Li-Reddy-2019,salerno2023high,Wiegrebe:2024aa,chen2024introduction}, with dedicated reviews of deep survival methods \cite{chen2024introduction} and practical introductions to the field \cite{EmmertStreib-Dehmer-19}.

The Cox proportional hazards model \cite{Cox-1972} is the most widely adopted classical approach. It relies on the proportional hazards assumption, which may not hold in many applications \cite{grambsch1994proportional}. Kaplan-Meier estimation \cite{Kaplan-Meier-58} remains the gold standard for nonparametric survival curve estimation, providing empirical estimates without distributional assumptions. Many machine learning survival models build upon the Cox  model, primarily by relaxing its linear relationship assumption \cite{Widodo-Yang-2011,Witten-Tibshirani-2010}. However, most of these extensions retain some of the fundamental limitations of the original Cox model, such as the proportional hazards assumption or semi-parametric structure. 
A significant portion of modern survival models extend classical machine learning approaches to handle censored data. Random Survival Forests \cite{Ishwaran-etal-2008} extend random forests to survival analysis and provide greater flexibility than parametric models.  Gradient boosted survival models \cite{ridgeway1999state} similarly extend the corresponding classical machine learning model. 

Neural network-based survival models have proliferated, including DeepSurv \cite{Katzman-etal-2018}, which extends Cox regression, and various other architectures designed specifically for survival prediction \cite{chen2024introduction,ren2019deep,Steingrimsson-Morrison-20, Tarkhan-etal-21,Zhong-Mueller-Wang-21}. More recently, transformer-based architectures have been adapted for survival analysis~\cite{hu2021transformer,Li-Zhu-Yao-Huang-22,Lv-Lin-etal-22, Shen-liu-etal-22,tang2023explainable,Wang-Sun-22}, leveraging attention mechanisms for improved feature representation. Attention-based deep survival models have also been independently proposed \cite{Li-Krivtsov-Arora-22,Sun-Dong-etal-21}, while tree-based methods like random survival forests \cite{Wright-etal-2017} extend ensemble approaches to survival settings.

\textbf{Generative Models and Diffusion Models}
Generative models have been increasingly applied to healthcare. VAEs have been explored for survival analysis \cite{apellaniz2024leveraging}, while GAN-based approaches \cite{chapfuwa2018adversarial} learn to generate synthetic survival data but suffer from training instability.
Diffusion probabilistic models have achieved state-of-the-art results in different applications \cite{dhariwal2021diffusion,Ho2020ddpm,kotelnikov2023tabddpm,pmlr-v139-nichol21a,peebles2023dit}. In the tabular domain, models such as TabDDPM \cite{kotelnikov2023tabddpm} and TabDiff \cite{shi2025tabdiff} adapt diffusion processes to mixed continuous and categorical variables, providing strong generative baselines for tabular data. Nevertheless, these models are not specifically designed for survival prediction and do not directly address the problem of estimating individual conditional survival functions from censored observations.

An interesting diffusion model for generating synthetic Data, called SurvDiff, is proposed in \cite{brockschmidt2025survdiff}. This is  an end-to-end diffusion model for generating synthetic survival datasets. SurvDiff jointly generates mixed-type covariates, observed times, and censoring indicators, and introduces a survival-tailored loss to preserve event-time structure and censoring mechanisms. Its goal is to approximate the full data-generating distribution of survival datasets so that synthetic samples can be used for downstream survival analysis without direct access to the original patient-level data. This objective is fundamentally different from the goal of the present work. SDPM does not aim to synthesize complete patient records or model the marginal distribution of covariates. Instead, it treats the feature vector $\mathbf{x}$ as given and models the conditional distribution of the observable survival outcome, $\mathbb{P}(T,\delta \mid \mathbf{x})$, in the target space of observed times and censoring indicators. Thus, while SurvDiff is a generative model for synthetic survival data generation in the joint space of covariates and outcomes, SDPM is a conditional generative model for survival regression, where diffusion is used to generate outcome samples for a fixed object and the survival function is reconstructed from these samples using the Kaplan--Meier estimator.

\section{Introduction to survival analysis}

In survival analysis, one considers objects (e.g., patients or technical devices) characterized by feature vectors $\mathbf{x}$ for which a terminal event may occur during an observation period. In most practical settings, the observation period is right-bounded; in this case, censoring is referred to as right censoring. If the event occurs during the observation period, the corresponding observation is called uncensored and has censoring indicator $\delta = 1$; otherwise, the observation is censored and $\delta = 0$. Let $E$ denote the random event time and $C$ the censoring time. Then the dataset $\mathcal{D}$ contains the observed time $T = \min\{E, C\}$ and is represented by triples $\{\mathbf{x}_i, t_i, \delta_i\}_{i \in \mathcal{I}}$.

The main characteristic of an object in survival analysis is the survival function
\begin{equation}\label{1}
S(t \mid \mathbf{x}) \equiv \mathbb{P}\{E > t \mid \mathbf{x}\}.
\end{equation}

The goal of survival models is to estimate this function, since other quantities of interest can be derived from it. For example, the expected event time can be expressed as
\begin{equation}\label{time_expectation}
    \mathbb{E}[E \mid \mathbf{x}] = \int_{0}^{\infty} S(t \mid \mathbf{x}) dt.
\end{equation}

Under right censoring, only the quantity $T = \min\{E, C\}$ and the indicator $\delta$ are observed, while the true event time $E$ remains unknown for censored observations. This means that the conditional distribution of event times is not fully observed, since for censored objects we only know that $E > C$.

In this work, we consider the problem of estimating the conditional joint distribution of the observed time and censoring indicator, $\mathbb{P}(T, \delta \mid \mathbf{x})$. Unlike classical approaches, we directly model this distribution using a generative model, which makes it possible to obtain a sample set
\begin{equation}\label{2}
\mathcal{T}(\mathbf{x}) = \{(t_i(\mathbf{x}), \delta_i(\mathbf{x}))\}_{i=1}^{K}
\end{equation}
for a given feature vector. This sample set is then used to construct the survival estimate $\hat{S}(t \;|\; \mathbf{x})$ via the Kaplan-Meier estimator. Since the generated values depend on the feature vector $\mathbf{x}$, the resulting survival function is conditional, even though it is obtained using a nonparametric method. Thus, survival function estimation is reduced to generating samples from a conditional distribution, which enables the use of nonparametric methods without introducing additional assumptions about the shape of the distribution.

\section{Survival Diffusion Probabilistic Model}

Unlike the standard use of diffusion models, where they are employed to generate new objects in feature space \cite{kotelnikov2023tabddpm}, in this work we propose a different approach: modeling the response distribution. Using a diffusion process, for a given feature vector $\mathbf{x}$ we generate a new conditional sample of observed times and censoring indicators, $\mathcal{T}(\mathbf{x})$. In contrast to other survival analysis models, SDPM operates in continuous time and is not tied to any discretization grid. Nevertheless, for fair comparison with other methods, a stepwise survival estimate $\hat{S}(t \mid \mathbf{x})$ is constructed from the sample set $\mathcal{T}(\mathbf{x})$ using the nonparametric Kaplan-Meier estimator on a discrete grid formed by the unique event times in the training set.

To improve convergence and numerical stability, SDPM operates in a transformed space of times $T$ and labels $\delta$. Event times in real datasets are often on the scale of hundreds or thousands, which complicates neural network modeling. For this reason, the diffusion process is applied to transformed targets
\begin{equation}\label{3}
\tilde{t} = \frac{\ln(t)-\mu}{\sigma},
\end{equation}
where $\mu$ and $\sigma$ are the mean and standard deviation of $\ln(t)$. The discrete censoring labels $\delta$ are replaced by continuous random variables: $\delta = 0$ is mapped to $\tilde{\delta}_0 \sim \mathcal{N}(-1, 0.25)$, while $\delta = 1$ is mapped to $\tilde{\delta}_1 \sim \mathcal{N}(1, 0.25)$. Thus, in the latent space the censoring label is represented as a mixture of two normal distributions corresponding to $\delta = 0$ and $\delta = 1$. For each training mini-batch, the latent censoring values $\tilde{\delta}$ are resampled independently for all observations according to their original binary censoring labels. In this way, the time transformation is invertible, while the original binary censoring label can be recovered from the sign of $\tilde{\delta}$. To recover $t$ from $\tilde{t}$, one applies the inverse standardization followed by the exponential transform:
\begin{equation}\label{4}
t = \exp(\sigma \tilde{t} + \mu).
\end{equation}

To recover the original discrete censoring label, it is sufficient to determine the sign of $\tilde{\delta}$:
\begin{equation}\label{5}
\delta = \mathbf{1}\big(\tilde{\delta} > 0\big).
\end{equation}

Denoising Diffusion Probabilistic Models (DDPM) \cite{Ho2020ddpm} define a family of latent variables $\{\tau_i\}_{i=0}^{r}$, where $\tau_0$ corresponds to the original data and $\tau_r$ corresponds to noise distributed according to the standard normal distribution. In our setting, the latent variable is a two-dimensional vector,
\begin{equation}\label{6}
\tau_i = (\tilde{t}_i, \tilde{\delta}_i).
\end{equation}

The forward diffusion process is defined as a Markov chain that progressively adds Gaussian noise to the data:
\begin{equation}\label{7}
q(\tau_i \mid \tau_{i-1}) = \mathcal{N}\big(\tau_i; \sqrt{1-\beta_i}\,\tau_{i-1}, \beta_i \mathbf{I}\big),
\end{equation}
where $\{\beta_i\}_{i=1}^{r}$ is a predefined variance schedule. In this work, we use the cosine schedule \cite{pmlr-v139-nichol21a}. By composing the forward transitions, one can derive an explicit expression for $q(\tau_i \mid \tau_0)$, which allows $\tau_i$ to be sampled directly from $\tau_0$ without simulating all intermediate steps.

The reverse process aims to reconstruct the data from noise and is also modeled as a Markov chain with parameterized transitions:
\begin{equation}
p_\theta(\tau_{i-1} \mid \tau_i, \mathbf{x}) =
\mathcal{N}\!\left(
\tau_{i-1};
\boldsymbol{\mu}_\theta(\tau_i, i, \mathbf{x}),
\boldsymbol{\Sigma}_\theta(\tau_i, i, \mathbf{x})
\right).
\end{equation}
where the distribution parameters are predicted by a neural network with parameters $\theta$. In practice, the model is trained not to predict $\boldsymbol{\mu}_\theta$ directly, but to reconstruct the Gaussian noise $\boldsymbol{\varepsilon}$ added during the forward process, which leads to the following simple mean squared error objective:
\begin{equation}
\mathcal{L}(\theta) =
\mathbb{E}_{\tau_0,\,\varepsilon,\,i,\,\mathbf{x}}
\left[
\left\|
\varepsilon - \varepsilon_\theta(\tau_i, i, \mathbf{x})
\right\|^2
\right].
\end{equation}

\begin{figure}[H]
    \centering
    \includegraphics[width=\linewidth]{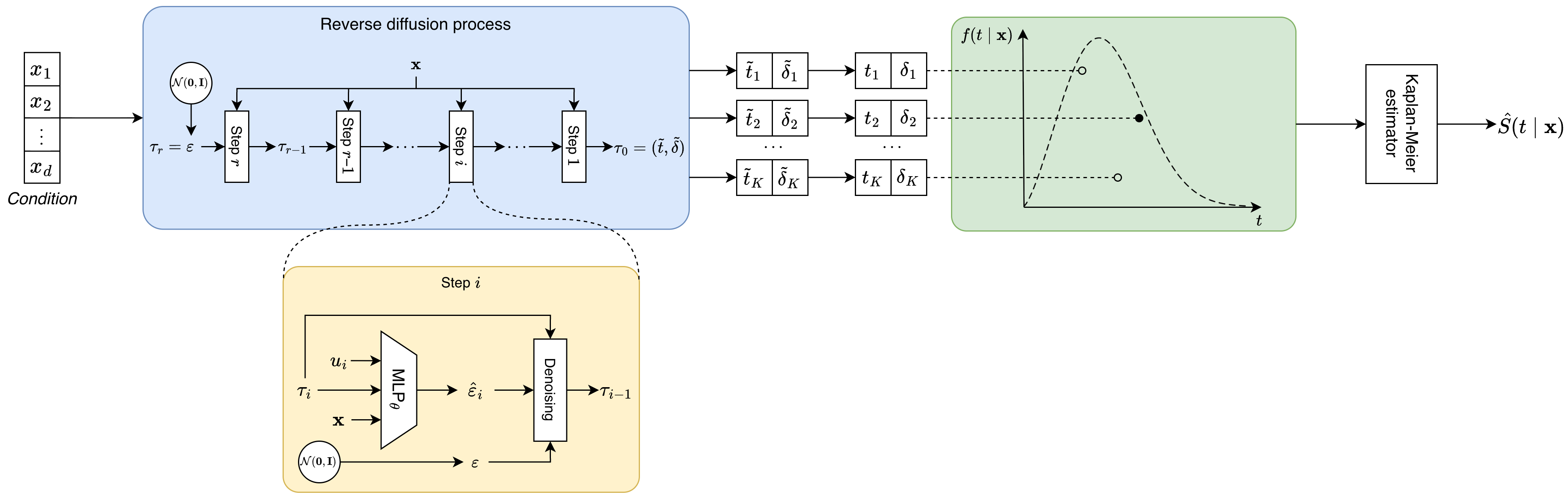}
    \caption{Schematic illustration of the SDPM generation pipeline
    }
    \label{fig:sdpm_generation}
\end{figure}

After training, samples are generated by sequentially applying the reverse process starting from $\tau_r \sim \mathcal{N}(\mathbf{0}, \mathbf{I})$ and ending at $\tau_0$, which is interpreted as an event realization (either observed or censored), $(t, \delta)$, for the feature vector $\mathbf{x}$. In particular, given the current latent variable $\tau_i$, the corresponding previous state $\tau_{i-1}$ is computed as:
\begin{equation}\label{denoising}
\tau_{i-1} = \frac{1}{\sqrt{\alpha_i}} \left( \tau_i - \frac{\beta_i}{\sqrt{1-\bar{\alpha}_i}} \, \varepsilon_\theta(\tau_i, i, \mathbf{x}) \right) + \sqrt{\beta_i} \, \varepsilon,
\end{equation}
where $\bar{\alpha}_i=\prod_{j=1}^i\alpha_j$, $\varepsilon \sim \mathcal{N}(\mathbf{0}, \mathbf{I})$, and $\varepsilon_\theta(\tau_i, i, \mathbf{x})$ is the neural network prediction (denoted as $\hat\varepsilon_{i}$ in Figure~\ref{fig:sdpm_generation}). This operation corresponds to the ``Denoising'' block in Figure~\ref{fig:sdpm_generation}.
To construct the set $\mathcal{T}(\mathbf{x})$, the reverse process is repeated $K$ times, after which the Kaplan-Meier estimator is applied to obtain $\hat{S}(t \mid \mathbf{x})$.

The neural network takes as input the feature vector $\mathbf{x}$, the diffusion-step embedding $u_i$, and the noisy variable $\tau_i$. To improve training stability, we optionally use Adaptive Layer Normalization with zero initialization (AdaLN-Zero), following \cite{peebles2023dit}, with the specific configuration selected using Optuna \cite{akiba2019optuna}. In addition, survival datasets, particularly in medical applications, often contain many categorical variables and only a small number of continuous features, which can be challenging for neural networks. Therefore, categorical features are processed using trainable embeddings, while continuous features are mapped to trainable Fourier features \cite{tancik2020ffn}. Specifically, for each continuous feature $x_i$, we use
\begin{equation}
    \gamma(x_i) =
    \big[
    \sin(2\pi w_{i1}x_i),\cos(2\pi w_{i1}x_i), \dots,\sin(2\pi w_{im}x_i),\cos(2\pi w_{im}x_i)
    \big],
\end{equation}
where $\{w_{ij}\}_{j=1}^{m}$ are trainable frequency coefficients initialized from a normal distribution. $m$ is a hyperparameter.

To illustrate the inference pipeline of the proposed model, we provide a schematic overview in Figure~\ref{fig:sdpm_generation}. The upper part of the figure shows the full reverse diffusion process, where the latent variable is iteratively transformed from Gaussian noise $\tau_r \sim \mathcal{N}(\mathbf{0}, \mathbf{I})$ to $\tau_0 = (\tilde{t}, \tilde{\delta})$ conditioned on the feature vector $\mathbf{x}$. 

The lower part details a single reverse diffusion step. At each step $i$, the neural network predicts the noise component $\varepsilon_\theta(\tau_i, i, \mathbf{x})$ (denoted as $\hat\varepsilon_{i}$), which is then used in the denoising operation, according to Equation~\eqref{denoising}, to obtain the previous latent state $\tau_{i-1}$.

By repeating the reverse process $K$ times, a set of samples $\mathcal{T}(\mathbf{x})$ is obtained. These samples are mapped to $(t, \delta)$ pairs and used to estimate the survival function via the Kaplan-Meier estimator.

\section{Experiments}

The experiments were conducted on real-world datasets collected from open sources. The experimental setup includes six survival models and ten datasets. The methods are compared using three metrics: C-index, integrated time-dependent AUC, and IBS.

\subsection{Data}

All datasets were preprocessed in the same way for all considered models. Numerical features were standardized, while categorical features were encoded using one-hot encoding. Missing numerical values were imputed with the corresponding feature means, and missing categorical values were replaced with an additional category. Samples for which all features were missing were removed. The ten datasets used in the experiments and their main characteristics after preprocessing are presented in Table~\ref{tab:datasets}.

\begin{table}[H]
    \caption{List of the datasets used in the experiments}
    \centering
    \begin{tabular}{lcccc}
        \toprule
        Name & Number of samples & Numeric & Categorical & Event rate \\
        \midrule
        FLC \cite{dispenzieri2012use} & 7874 & 4 & 34 & 28\%\\
        Ovarian \cite{ganzfried2013database} & 912 & 158 & 6 & 60\%\\
        PBC \cite{fleming2013counting} & 418 & 11 & 15 & 39\%\\
        Retinopathy \cite{blair19805} & 394 & 2 & 14 & 39\%\\
        Rotterdam \cite{royston2013external} & 2982 & 5 & 6 & 43\%\\
        SEER & 4024 & 4 & 21 & 15\%\\
        SUPPORT \cite{connors1995controlled} & 9105 & 37 & 30 & 68\%\\
        TCGA-GBM \cite{liu2018integrated} & 596 & 5 & 20 & 82\%\\
        VLBW \cite{oshea1992} & 617 & 7 & 49 & 17\% \\ 
        WHAS500 \cite{hosmer2008applied} & 500 & 6 & 8 & 43\%\\
        \bottomrule
    \end{tabular}
    \label{tab:datasets}
\end{table}

The SEER dataset was obtained from the open-source repository available at \url{https://dx.doi.org/10.21227/a9qy-ph35}. The TCGA-GBM dataset was extracted from the TCGA PanCanAtlas dataset \cite{liu2018integrated} by selecting patients diagnosed with Glioblastoma Multiforme (GBM). The Ovarian dataset is a subset of four studies from the curatedOvarianData collection \cite{ganzfried2013database}. It includes 158 gene expression features, whereas all other datasets contain only clinical variables.

\subsection{Models}

The numerical experiments compare six survival models:

\begin{itemize}
    \item Random Survival Forest \cite{Ishwaran-etal-2008} -- a nonparametric ensemble method for survival analysis that serves as a strong tabular baseline due to its ability to model nonlinearities and feature interactions without explicit distributional assumptions.
    \item DeepSurv \cite{Katzman-etal-2018} -- a neural-network version of the Cox proportional hazards model, used as a standard deep learning baseline for capturing nonlinear feature effects while retaining the classical problem formulation.
    \item DeepHit \cite{lee2018deephit} -- a neural survival model that directly approximates the event-time distribution, allowing it to capture more complex dependencies and serving as a representative modern method without the proportional hazards assumption.
    \item XGBSEKaplanNeighbors \cite{xgbse2020github} -- a boosting-based approach combining gradient boosting with Kaplan-Meier estimation, acting as a strong tabular baseline.
    \item XGBSEStackedWeibull \cite{xgbse2020github} -- a gradient boosting model followed by parametric Weibull approximation, representing a hybrid between nonparametric and parametric survival approaches.
    \item SDPM -- the deep learning model proposed in this work, which models the distribution of survival outcomes using a diffusion process.
\end{itemize}

For brevity, XGBSEKaplanNeighbors and XGBSEStackedWeibull are referred to as GBM-KM and GBM-Weibull, respectively, in Tables~\ref{tab:res_c_index}--\ref{tab:res_ibs} and Figures~\ref{fig:ranks_c_index}--\ref{fig:ranks_ibs}.

\subsection{Methodology}

To ensure an objective comparison, we used 4-fold cross-validation. Within each fold, hyperparameters were optimized using Optuna \cite{akiba2019optuna} for 100 trials. The hyperparameter search spaces for all models are provided in Appendix~\ref{sec:hyperparams}. A validation subset containing 25\% of the current training fold was used for hyperparameter selection. For methods supporting early stopping, an additional validation subset of size 25\% was sampled from the current training data for that purpose. All splits were generated using stratification with respect to the proportion of censored observations. Each cross-validation experiment was repeated independently 10 times to obtain reliable statistics.

The quality of survival predictions is evaluated using three metrics: Harrell's C-index \cite{harrell1996multivariable}, time-dependent area under the ROC curve (time-dependent AUC) \cite{uno2007evaluating}, and the integrated Brier score (IBS) \cite{graf1999assessment}. The C-index evaluates the correctness of ranking the expected event times according to the observed times in the test set:
\begin{equation}\label{c_index}
    C = \frac{\sum\limits_{i,j} \mathbf{1}\big(t_i < t_j\big)\mathbf{1}\big(\mathbb{E}[E \mid \mathbf{x}_i] < \mathbb{E}[E \mid \mathbf{x}_j]\big)\delta_i}{\sum\limits_{i,j} \mathbf{1}\big(t_i < t_j\big)\delta_i }.
\end{equation}

A value of 0.5 corresponds to random ranking, while larger values indicate better ordering of objects with respect to event risk. 

The time-dependent AUC evaluates the discriminative ability of the model at a specific time point $t$. Following the cumulative/dynamic definition \cite{uno2007evaluating}, it is defined as
\begin{equation}\label{td_auc}
    \widetilde{\mathrm{AUC}}(t) =
    \mathbb{P}\left\{
    \hat{S}(t \mid \mathbf{x}_i)
    <
    \hat{S}(t \mid \mathbf{x}_j)
    \,\middle|\,
    t_i \le t,\ t_j > t
    \right\},
\end{equation}
where individuals experiencing the event before time $t$ are treated as cases, while individuals surviving beyond $t$ are treated as controls. To obtain a scalar summary measure, the time-dependent AUC is integrated over time using the Kaplan-Meier estimate of the survival function:
\begin{equation}\label{i_auc}
    \mathrm{AUC} =
    \frac{1}{\hat{S}_{\mathrm{KM}}(t_{\mathrm{min}}) - \hat{S}_{\mathrm{KM}}(t_{\mathrm{max}})}
    \int_{t_{\mathrm{min}}}^{t_{\mathrm{max}}}
    \widetilde{\mathrm{AUC}}(t)\, d\hat{S}_{\mathrm{KM}}(t),
\end{equation}
where $\hat{S}_{\mathrm{KM}}(t)$ denotes the Kaplan-Meier estimate of the survival function. Larger values of the AUC correspond to better discriminative performance across time.

IBS, in contrast, characterizes the accuracy of the estimated survival function. Unlike the C-index and time-dependent AUC, smaller values correspond to better predictions:
\begin{equation}\label{ibs}
    \mathrm{IBS} = \frac{1}{t_{\max}} \int_{0}^{t_{\max}} \mathrm{BS}(t)\, dt,\ \text{where}
\end{equation}
\begin{equation}
    \mathrm{BS}(t) = \frac{1}{n} \sum_{i=1}^{n} \left[
    \frac{\mathbf{1}(t_i \le t)\delta_i}{\hat{G}(t_i)} (0 - \hat{S}(t \mid \mathbf{x}_i))^2
    +
    \frac{\mathbf{1}(t_i > t)}{\hat{G}(t)} (1 - \hat{S}(t \mid \mathbf{x}_i))^2
    \right],
\end{equation}
where $\hat{G}(t)$ is the Kaplan-Meier estimate of the censoring survival function $\mathbb{P}\{C > t\}$.

To ensure fair comparison across methods, all metrics are evaluated on the same discrete time grid $(\overline{t}_0, \overline{t}_1, \ldots, \overline{t}_n)$, where $\overline{t}_0 \equiv 0$, obtained from the unique event times in the training set. In this case, the survival function takes a stepwise form:
\begin{equation}
S(t \mid \mathbf{x}) = S_i(\mathbf{x}), \quad t \in [\overline{t}_{i-1}, \overline{t}_i).
\end{equation}

\subsection{Results}

The results averaged over all experimental runs are presented in Tables~\ref{tab:res_c_index}, \ref{tab:res_auc}, and~\ref{tab:res_ibs} for C-index, time-dependent AUC, and IBS, respectively. The best result for each dataset is highlighted in bold.

\begin{table}[H]
    \caption{Numerical results, C-index ($\uparrow$)}
      \centering
    {
    \begin{tabular}{lcccccc}
\toprule
 & DeepHit & DeepSurv & GBM-KM & GBM-Weibull & RSF & SDPM \\
\midrule
FLC & $0.930 {\scriptstyle \pm 0.001}$ & $0.934 {\scriptstyle \pm 0.001}$ & $0.931 {\scriptstyle \pm 0.001}$ & $0.907 {\scriptstyle \pm 0.046}$ & $0.935 {\scriptstyle \pm 0.001}$ & $\mathbf{0.936 {\scriptstyle \pm 0.001}}$ \\
Ovarian & $\mathbf{0.632 {\scriptstyle \pm 0.006}}$ & $0.629 {\scriptstyle \pm 0.010}$ & $0.617 {\scriptstyle \pm 0.012}$ & $0.621 {\scriptstyle \pm 0.028}$ & $0.624 {\scriptstyle \pm 0.008}$ & $0.625 {\scriptstyle \pm 0.016}$ \\
PBC & $0.957 {\scriptstyle \pm 0.008}$ & $0.983 {\scriptstyle \pm 0.003}$ & $0.955 {\scriptstyle \pm 0.053}$ & $0.956 {\scriptstyle \pm 0.056}$ & $0.963 {\scriptstyle \pm 0.004}$ & $\mathbf{0.993 {\scriptstyle \pm 0.004}}$ \\
Retinopathy & $0.600 {\scriptstyle \pm 0.016}$ & $0.617 {\scriptstyle \pm 0.019}$ & $0.622 {\scriptstyle \pm 0.024}$ & $0.614 {\scriptstyle \pm 0.038}$ & $\mathbf{0.632 {\scriptstyle \pm 0.016}}$ & $0.621 {\scriptstyle \pm 0.019}$ \\
Rotterdam & $0.699 {\scriptstyle \pm 0.003}$ & $0.705 {\scriptstyle \pm 0.004}$ & $0.693 {\scriptstyle \pm 0.017}$ & $0.702 {\scriptstyle \pm 0.028}$ & $\mathbf{0.707 {\scriptstyle \pm 0.001}}$ & $0.705 {\scriptstyle \pm 0.003}$ \\
SEER & $0.728 {\scriptstyle \pm 0.006}$ & $0.712 {\scriptstyle \pm 0.033}$ & $0.712 {\scriptstyle \pm 0.009}$ & $0.727 {\scriptstyle \pm 0.005}$ & $0.725 {\scriptstyle \pm 0.003}$ & $\mathbf{0.729 {\scriptstyle \pm 0.006}}$ \\
SUPPORT & $0.883 {\scriptstyle \pm 0.003}$ & $0.890 {\scriptstyle \pm 0.001}$ & $0.892 {\scriptstyle \pm 0.001}$ & $\mathbf{0.894 {\scriptstyle \pm 0.000}}$ & $0.857 {\scriptstyle \pm 0.001}$ & $0.892 {\scriptstyle \pm 0.002}$ \\
TCGA-GBM & $0.874 {\scriptstyle \pm 0.006}$ & $0.852 {\scriptstyle \pm 0.016}$ & $0.881 {\scriptstyle \pm 0.005}$ & $0.873 {\scriptstyle \pm 0.028}$ & $0.851 {\scriptstyle \pm 0.004}$ & $\mathbf{0.883 {\scriptstyle \pm 0.004}}$ \\
VLBW & $0.782 {\scriptstyle \pm 0.067}$ & $\mathbf{0.893 {\scriptstyle \pm 0.012}}$ & $0.782 {\scriptstyle \pm 0.070}$ & $0.691 {\scriptstyle \pm 0.122}$ & $0.877 {\scriptstyle \pm 0.011}$ & $0.888 {\scriptstyle \pm 0.010}$ \\
WHAS500 & $0.754 {\scriptstyle \pm 0.007}$ & $0.758 {\scriptstyle \pm 0.009}$ & $0.751 {\scriptstyle \pm 0.015}$ & $0.704 {\scriptstyle \pm 0.078}$ & $\mathbf{0.767 {\scriptstyle \pm 0.007}}$ & $0.760 {\scriptstyle \pm 0.008}$ \\
\bottomrule
\end{tabular}

    }
    \label{tab:res_c_index}
\end{table}

\begin{table}[H]
\caption{Numerical results, integrated time-dependent AUC ($\uparrow$)}
     \centering
    {
    \begin{tabular}{lcccccc}
\toprule
 & DeepHit & DeepSurv & GBM-KM & GBM-Weibull & RSF & SDPM \\
\midrule
FLC & $0.947 {\scriptstyle \pm 0.002}$ & $0.951 {\scriptstyle \pm 0.002}$ & $0.938 {\scriptstyle \pm 0.003}$ & $0.899 {\scriptstyle \pm 0.075}$ & $0.953 {\scriptstyle \pm 0.001}$ & $\mathbf{0.954 {\scriptstyle \pm 0.001}}$ \\
Ovarian & $\mathbf{0.666 {\scriptstyle \pm 0.012}}$ & $0.665 {\scriptstyle \pm 0.015}$ & $0.661 {\scriptstyle \pm 0.016}$ & $0.645 {\scriptstyle \pm 0.036}$ & $0.664 {\scriptstyle \pm 0.012}$ & $0.659 {\scriptstyle \pm 0.024}$ \\
PBC & $0.976 {\scriptstyle \pm 0.007}$ & $0.994 {\scriptstyle \pm 0.002}$ & $0.957 {\scriptstyle \pm 0.052}$ & $0.957 {\scriptstyle \pm 0.057}$ & $0.973 {\scriptstyle \pm 0.007}$ & $\mathbf{0.998 {\scriptstyle \pm 0.002}}$ \\
Retinopathy & $0.611 {\scriptstyle \pm 0.029}$ & $0.645 {\scriptstyle \pm 0.026}$ & $0.636 {\scriptstyle \pm 0.032}$ & $0.643 {\scriptstyle \pm 0.048}$ & $0.654 {\scriptstyle \pm 0.025}$ & $\mathbf{0.655 {\scriptstyle \pm 0.031}}$ \\
Rotterdam & $0.749 {\scriptstyle \pm 0.010}$ & $0.745 {\scriptstyle \pm 0.011}$ & $0.733 {\scriptstyle \pm 0.020}$ & $0.695 {\scriptstyle \pm 0.053}$ & $0.754 {\scriptstyle \pm 0.010}$ & $\mathbf{0.756 {\scriptstyle \pm 0.010}}$ \\
SEER & $\mathbf{0.742 {\scriptstyle \pm 0.006}}$ & $0.726 {\scriptstyle \pm 0.033}$ & $0.713 {\scriptstyle \pm 0.011}$ & $0.717 {\scriptstyle \pm 0.037}$ & $0.737 {\scriptstyle \pm 0.004}$ & $0.740 {\scriptstyle \pm 0.007}$ \\
SUPPORT & $0.945 {\scriptstyle \pm 0.003}$ & $0.947 {\scriptstyle \pm 0.002}$ & $0.948 {\scriptstyle \pm 0.002}$ & $0.932 {\scriptstyle \pm 0.058}$ & $0.940 {\scriptstyle \pm 0.003}$ & $\mathbf{0.951 {\scriptstyle \pm 0.001}}$ \\
TCGA-GBM & $0.929 {\scriptstyle \pm 0.006}$ & $0.906 {\scriptstyle \pm 0.014}$ & $\mathbf{0.935 {\scriptstyle \pm 0.004}}$ & $0.871 {\scriptstyle \pm 0.057}$ & $0.929 {\scriptstyle \pm 0.005}$ & $0.931 {\scriptstyle \pm 0.005}$ \\
VLBW & $0.867 {\scriptstyle \pm 0.026}$ & $0.879 {\scriptstyle \pm 0.018}$ & $0.854 {\scriptstyle \pm 0.026}$ & $0.830 {\scriptstyle \pm 0.079}$ & $\mathbf{0.900 {\scriptstyle \pm 0.014}}$ & $0.892 {\scriptstyle \pm 0.009}$ \\
WHAS500 & $0.784 {\scriptstyle \pm 0.007}$ & $0.778 {\scriptstyle \pm 0.014}$ & $0.775 {\scriptstyle \pm 0.015}$ & $0.741 {\scriptstyle \pm 0.055}$ & $\mathbf{0.802 {\scriptstyle \pm 0.011}}$ & $0.788 {\scriptstyle \pm 0.009}$ \\
\bottomrule
\end{tabular}

    }
    \label{tab:res_auc}
\end{table}


\begin{table}[H]
   \caption{Numerical results, IBS ($\downarrow$)}
    \centering
    {
    \begin{tabular}{lcccccc}
\toprule
 & DeepHit & DeepSurv & GBM-KM & GBM-Weibull & RSF & SDPM \\
\midrule
FLC & $0.074 {\scriptstyle \pm 0.012}$ & $0.046 {\scriptstyle \pm 0.001}$ & $0.048 {\scriptstyle \pm 0.001}$ & $0.073 {\scriptstyle \pm 0.013}$ & $0.046 {\scriptstyle \pm 0.001}$ & $\mathbf{0.046 {\scriptstyle \pm 0.001}}$ \\
Ovarian & $0.146 {\scriptstyle \pm 0.007}$ & $0.148 {\scriptstyle \pm 0.004}$ & $0.162 {\scriptstyle \pm 0.023}$ & $0.166 {\scriptstyle \pm 0.008}$ & $\mathbf{0.145 {\scriptstyle \pm 0.002}}$ & $0.166 {\scriptstyle \pm 0.017}$ \\
PBC & $0.103 {\scriptstyle \pm 0.027}$ & $0.021 {\scriptstyle \pm 0.003}$ & $0.037 {\scriptstyle \pm 0.021}$ & $0.084 {\scriptstyle \pm 0.022}$ & $0.053 {\scriptstyle \pm 0.003}$ & $\mathbf{0.011 {\scriptstyle \pm 0.004}}$ \\
Retinopathy & $0.203 {\scriptstyle \pm 0.006}$ & $\mathbf{0.190 {\scriptstyle \pm 0.002}}$ & $0.193 {\scriptstyle \pm 0.004}$ & $0.216 {\scriptstyle \pm 0.065}$ & $0.193 {\scriptstyle \pm 0.011}$ & $0.196 {\scriptstyle \pm 0.007}$ \\
Rotterdam & $0.173 {\scriptstyle \pm 0.009}$ & $0.150 {\scriptstyle \pm 0.001}$ & $0.153 {\scriptstyle \pm 0.003}$ & $0.156 {\scriptstyle \pm 0.007}$ & $0.148 {\scriptstyle \pm 0.001}$ & $\mathbf{0.146 {\scriptstyle \pm 0.002}}$ \\
SEER & $0.082 {\scriptstyle \pm 0.003}$ & $0.080 {\scriptstyle \pm 0.002}$ & $0.082 {\scriptstyle \pm 0.001}$ & $0.084 {\scriptstyle \pm 0.002}$ & $0.081 {\scriptstyle \pm 0.001}$ & $\mathbf{0.080 {\scriptstyle \pm 0.001}}$ \\
SUPPORT & $0.115 {\scriptstyle \pm 0.006}$ & $0.103 {\scriptstyle \pm 0.001}$ & $0.104 {\scriptstyle \pm 0.001}$ & $0.137 {\scriptstyle \pm 0.012}$ & $0.108 {\scriptstyle \pm 0.001}$ & $\mathbf{0.098 {\scriptstyle \pm 0.002}}$ \\
TCGA-GBM & $0.087 {\scriptstyle \pm 0.007}$ & $0.071 {\scriptstyle \pm 0.005}$ & $0.057 {\scriptstyle \pm 0.003}$ & $0.092 {\scriptstyle \pm 0.011}$ & $0.067 {\scriptstyle \pm 0.003}$ & $\mathbf{0.053 {\scriptstyle \pm 0.002}}$ \\
VLBW & $0.102 {\scriptstyle \pm 0.011}$ & $0.090 {\scriptstyle \pm 0.007}$ & $0.100 {\scriptstyle \pm 0.008}$ & $0.140 {\scriptstyle \pm 0.072}$ & $0.088 {\scriptstyle \pm 0.003}$ & $\mathbf{0.084 {\scriptstyle \pm 0.005}}$ \\
WHAS500 & $0.184 {\scriptstyle \pm 0.012}$ & $\mathbf{0.159 {\scriptstyle \pm 0.007}}$ & $0.183 {\scriptstyle \pm 0.004}$ & $0.193 {\scriptstyle \pm 0.010}$ & $0.160 {\scriptstyle \pm 0.002}$ & $0.161 {\scriptstyle \pm 0.005}$ \\
\bottomrule
\end{tabular}

    }
    \label{tab:res_ibs}
\end{table}

\begin{figure}[H]
    \centering
    \includegraphics[width=0.9\linewidth]{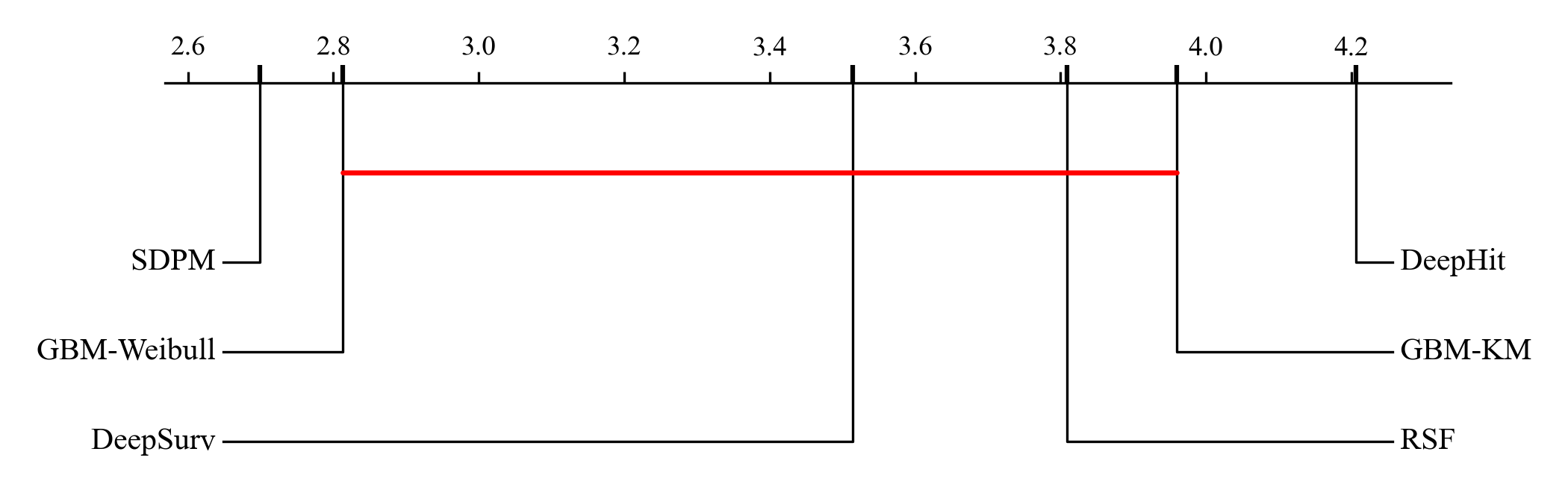}
    \caption{Critical difference diagram for C-index ranks ($\downarrow$)}
    \label{fig:ranks_c_index}
\end{figure}

\begin{figure}[H]
    \centering
    \includegraphics[width=0.9\linewidth]{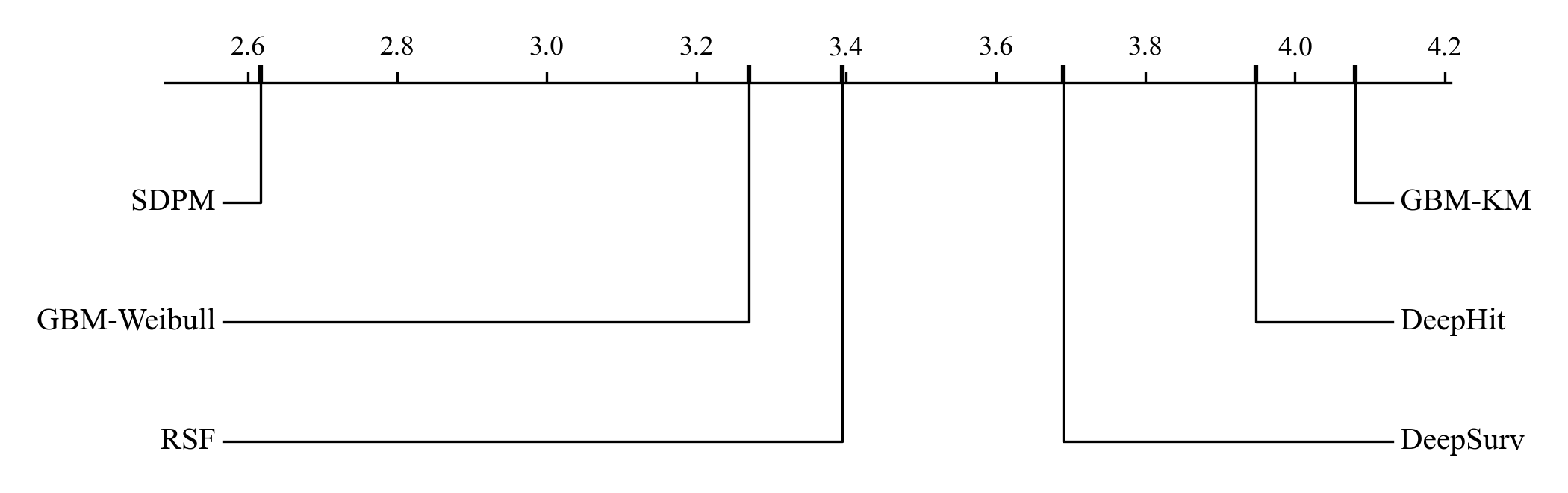}
    \caption{Critical difference diagram for AUC ranks ($\downarrow$)}
    \label{fig:ranks_auc}
\end{figure}

\begin{figure}[H]
    \centering
    \includegraphics[width=0.9\linewidth]{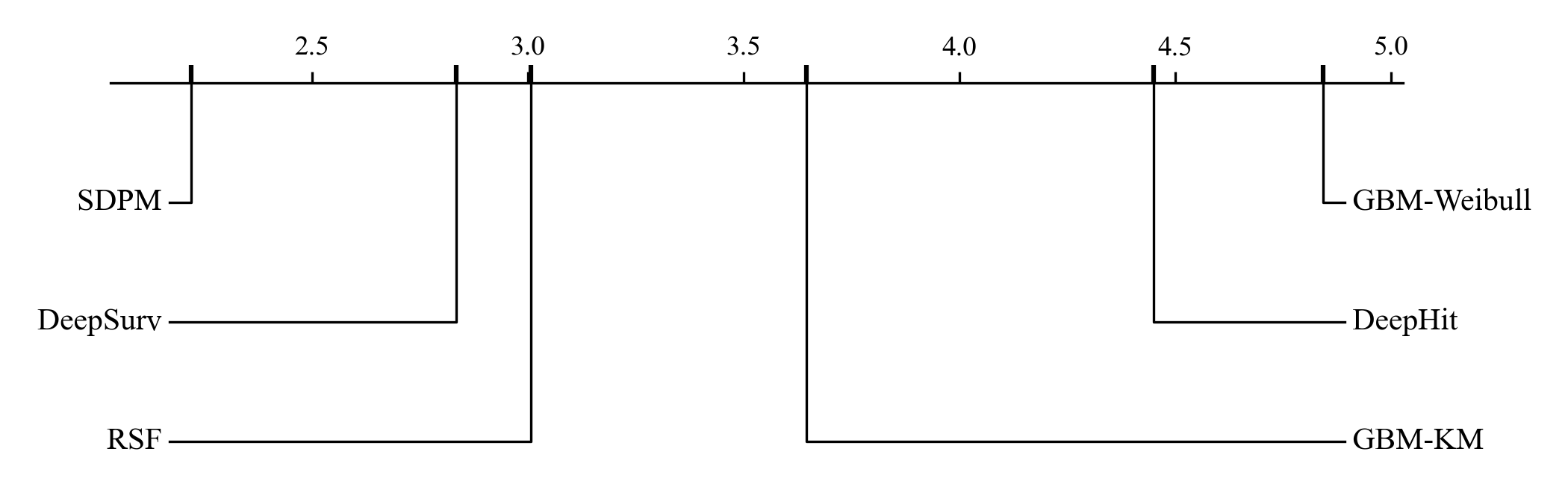}
    \caption{Critical difference diagram for IBS ranks ($\downarrow$)}
    \label{fig:ranks_ibs}
\end{figure}

Additionally, rank statistics were aggregated across all independent cross-validation folds and visualized using critical difference diagrams in Figures~\ref{fig:ranks_c_index}, \ref{fig:ranks_auc} and \ref{fig:ranks_ibs}. The ranks were computed taking into account the direction of each metric, so that lower ranks consistently correspond to better performance. Statistical significance of pairwise differences was assessed using the Wilcoxon signed-rank test, and differences were considered significant at the level $p < 0.05$.

\subsection{Analysis of results}

The results in Tables~\ref{tab:res_c_index}, \ref{tab:res_auc}, and~\ref{tab:res_ibs} show that the proposed SDPM method is competitive with established survival analysis approaches across all considered metrics. Overall, SDPM achieves strong performance both in terms of discriminative quality, measured by C-index and time-dependent AUC, and in terms of survival function calibration, measured by IBS.

For C-index, SDPM obtains the best result on four datasets: FLC, PBC, SEER, and TCGA-GBM. On several other datasets, its performance remains close to the best competing method. For example, on SUPPORT and VLBW, the difference from the best result is small, while on Rotterdam and WHAS500 SDPM remains comparable to the strongest baselines. This indicates that the proposed generative formulation is capable of producing survival estimates that preserve the correct ordering of objects by event risk.

The time-dependent AUC results provide a similar picture. SDPM achieves the best performance on five datasets, namely FLC, PBC, Retinopathy, Rotterdam, and SUPPORT, and remains close to the best method on several others. Since time-dependent AUC evaluates discrimination across the evaluation time horizon, these results suggest that SDPM provides competitive time-dependent risk separation, rather than only a good global ordering of expected event times.

The strongest advantage of SDPM is observed for IBS. The proposed method achieves the best IBS on seven out of ten datasets: FLC, PBC, Rotterdam, SEER, SUPPORT, TCGA-GBM, and VLBW. This is particularly important because IBS directly reflects the accuracy of the estimated survival function. The observed improvement is consistent with the main motivation of SDPM: by modeling the joint distribution of event times and censoring indicators, the method can recover not only a ranking of objects, but also a more accurate approximation of the survival distribution.


The critical difference diagrams in Figures~\ref{fig:ranks_c_index}, \ref{fig:ranks_auc}, and~\ref{fig:ranks_ibs} provide an aggregate view of the results across folds. In all three metrics, SDPM has the best average rank among the compared methods. The advantage is most pronounced for IBS, where SDPM is clearly separated from most competing approaches, supporting the conclusion that the proposed method is particularly effective for survival function estimation. For C-index and time-dependent AUC, the differences between the best-performing methods are smaller, which indicates that several methods achieve comparable discriminative performance.

Overall, the results show that SDPM provides a favorable balance between ranking quality and survival distribution estimation. While its discriminative performance is generally competitive with the strongest baselines, its main advantage lies in the quality of the estimated survival functions, as reflected by consistently strong IBS results and the best aggregate rank across the considered metrics.

\subsection{Additional experimental analysis}

\subsubsection{Influence of the number of generated samples}

Since in the proposed approach the survival function is estimated from a finite set of generated pairs $(T, \delta)$, the quality of the final estimate may depend on the number of samples $K$ used for Kaplan-Meier estimation. To verify this property, we conducted an additional study of the influence of $K$ on predictive quality and inference time.

In this experiment, the model was trained once with fixed hyperparameters on the training portion of the VLBW dataset. Then, for each value $K \in \{2^7,2^8,\ldots,2^{15}\}$, the sample set $\mathcal{T}(\mathbf{x})$ was repeatedly generated for the test objects, followed by survival estimation and metric computation. For each value of $K$, the procedure was repeated 100 times in order to capture the variability of the metrics induced by the stochasticity of the generation process.

\begin{figure}[H]
    \centering
    \includegraphics[width=0.9\linewidth]{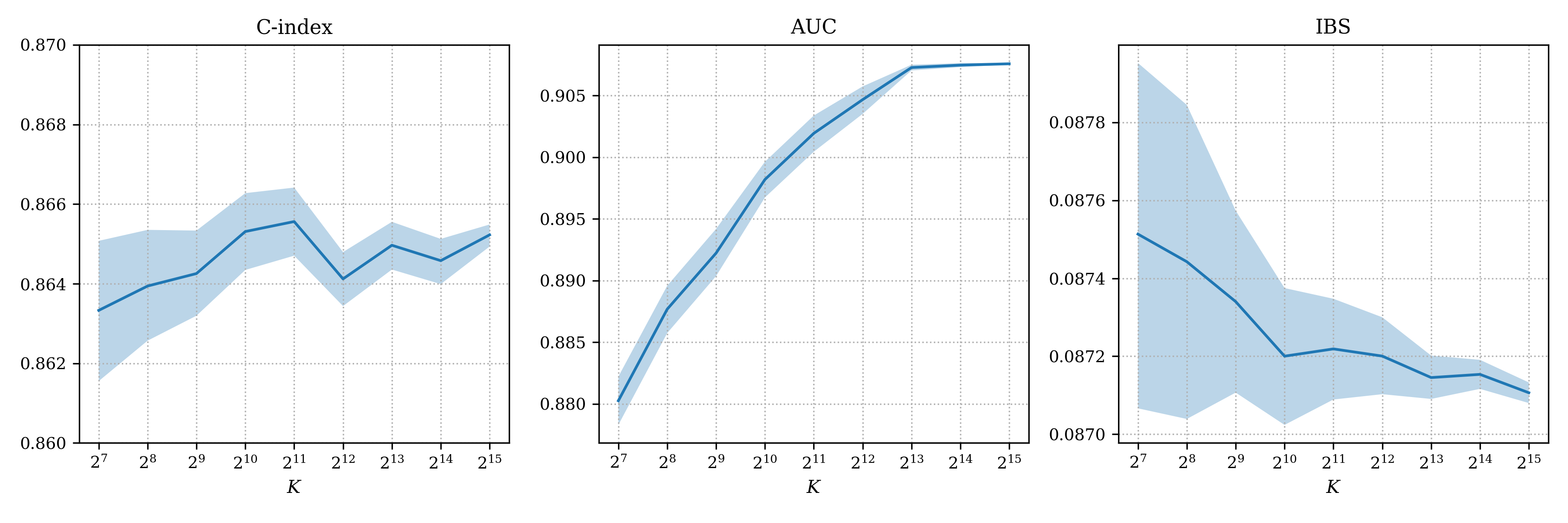}
    \caption{Influence of the number of generated $(T,\;\delta)$ pairs on C-index, integrated time-dependent AUC, and IBS on the VLBW dataset. Shaded regions correspond to 95\% confidence intervals estimated from 100 repetitions.}
    \label{fig:ablation_k}
\end{figure}

Figure~\ref{fig:ablation_k} presents the mean values of C-index, integrated time-dependent AUC, and IBS computed on the same discrete time grid, together with the corresponding 95\% confidence intervals estimated from 100 repetitions. The intervals were estimated by bootstrap resampling of the obtained metric values.

The results show that increasing the number of generated realizations mainly improves the stability and quality of distributional survival estimation. The most pronounced effect is observed for the integrated time-dependent AUC, which increases consistently as $K$ grows and approaches a plateau for large values of $K$. The IBS also decreases gradually, indicating a moderate improvement in survival probability calibration. In contrast, the C-index remains relatively stable across different values of $K$ and fluctuates only within a narrow range. This suggests that, for this dataset, the number of generated samples has a stronger influence on time-dependent discrimination and survival-function estimation than on the global ranking quality measured by the C-index.

\begin{figure}[H]
    \centering
    \includegraphics[width=0.7\linewidth]{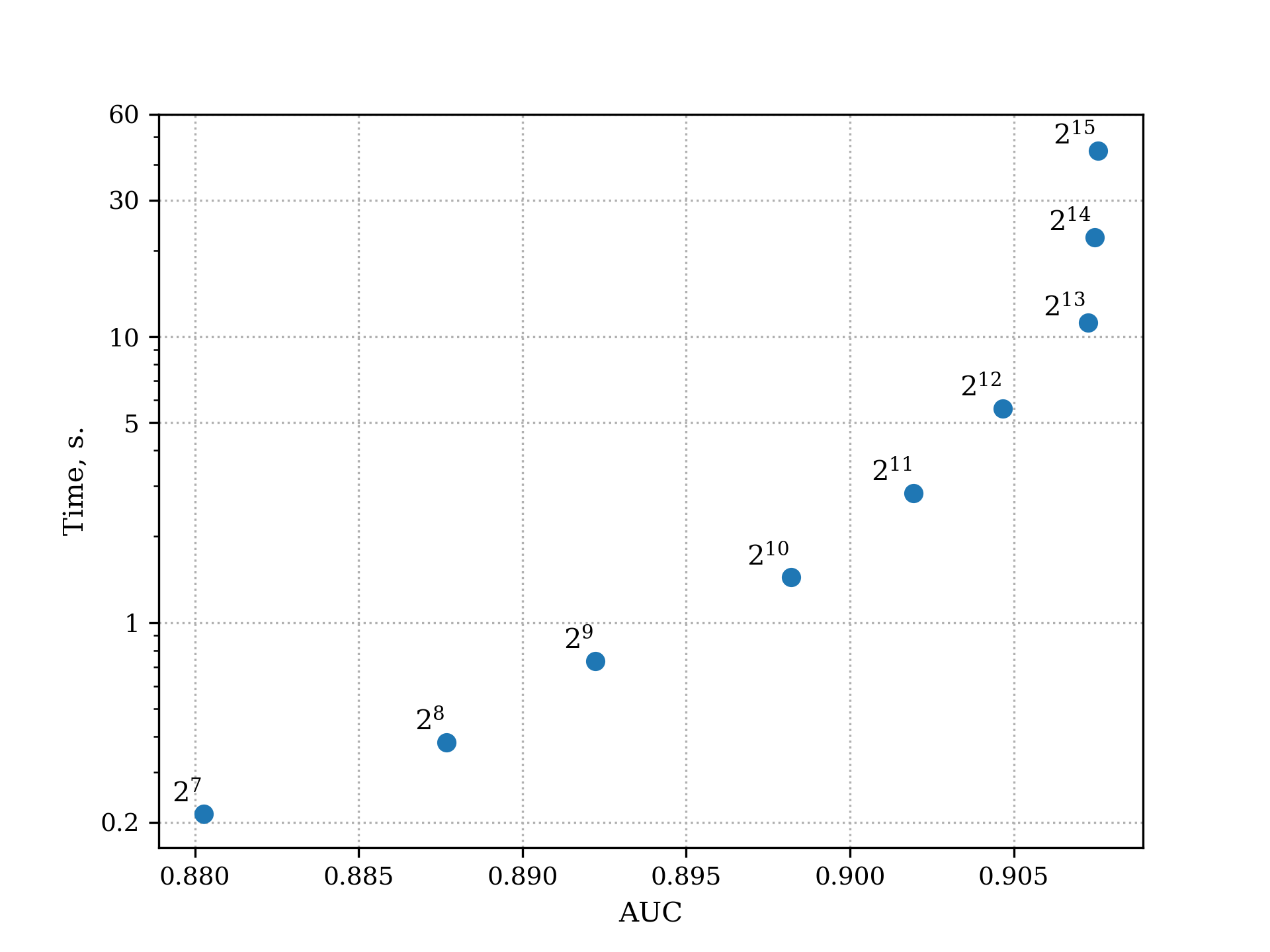}
    \caption{Trade-off between predictive quality, measured by mean integrated time-dependent AUC, and total generation time for different values of $K$ on the VLBW dataset. Each point is annotated with the corresponding value of $K=2^i$.}
    \label{fig:ablation_time}
\end{figure}

At the same time, increasing $K$ leads to a substantial increase in inference time, since each additional realization requires an additional reverse diffusion sampling trajectory. Figure~\ref{fig:ablation_time} illustrates the trade-off between the mean integrated time-dependent AUC and the total generation time for different values of $K$. The AUC increases steadily with $K$, while the total generation time grows rapidly, especially for larger sample sizes. This indicates that increasing $K$ can improve the quality of the survival estimate, but the resulting computational overhead becomes increasingly restrictive for extensive experimental studies.

Although the AUC continues to improve for larger values of $K$, the results do not indicate any degradation of predictive quality as the number of generated samples increases. This confirms that using more generated realizations generally leads to a more stable Kaplan-Meier reconstruction. However, the computational cost grows rapidly with $K$, making very large sample sizes impractical for large-scale benchmark experiments involving multiple datasets, repeated train-test splits, and hyperparameter configurations.

For this reason, we selected $K = 2^{11} = 2048$ for the main numerical experiments. This value provides a clear improvement over smaller sample sizes while keeping the inference time sufficiently low for extensive empirical evaluation. Larger values of $K$ may further improve time-dependent discrimination and calibration, but at a substantially higher computational cost.

\subsubsection{Influence of the number of diffusion steps}

In addition to the number of generated realizations $K$, an important design parameter of the proposed model is the number of reverse diffusion steps $r$. Since the reverse process is implemented as a discrete approximation of gradual denoising, changing $r$ may affect the quality of the generated sample set $\mathcal{T}(\mathbf{x})$ and, consequently, the resulting survival estimate.

To study this effect, we conducted an additional experiment on the TCGA-GBM dataset under the same fixed hyperparameter configuration as in the previous ablation study on $K$. The model was originally trained with $r = 20$ diffusion steps. After training, the learned neural network was kept fixed, while the number of reverse diffusion steps used during inference was varied as $r\in\{2^1,2^2,\ldots,2^7\}$.

This experiment is enabled by the functional parametrization of the diffusion-step encoding used in SDPM. The denoising network is conditioned on a sinusoidal encoding computed directly from the numerical step value. As a result, the trained network can be evaluated at step values that do not necessarily coincide with the original training discretization. For each tested value of $r$, we constructed the corresponding cosine variance schedule and linearly rescaled the new reverse-process step indices to the same numerical range of step values used during training. The generated samples were then used to estimate the survival function on the test subset, and the resulting C-index, integrated time-dependent AUC, and IBS values were computed over 100 independent repetitions with different random seeds.

\begin{figure}[H]
    \centering
    \includegraphics[width=0.9\linewidth]{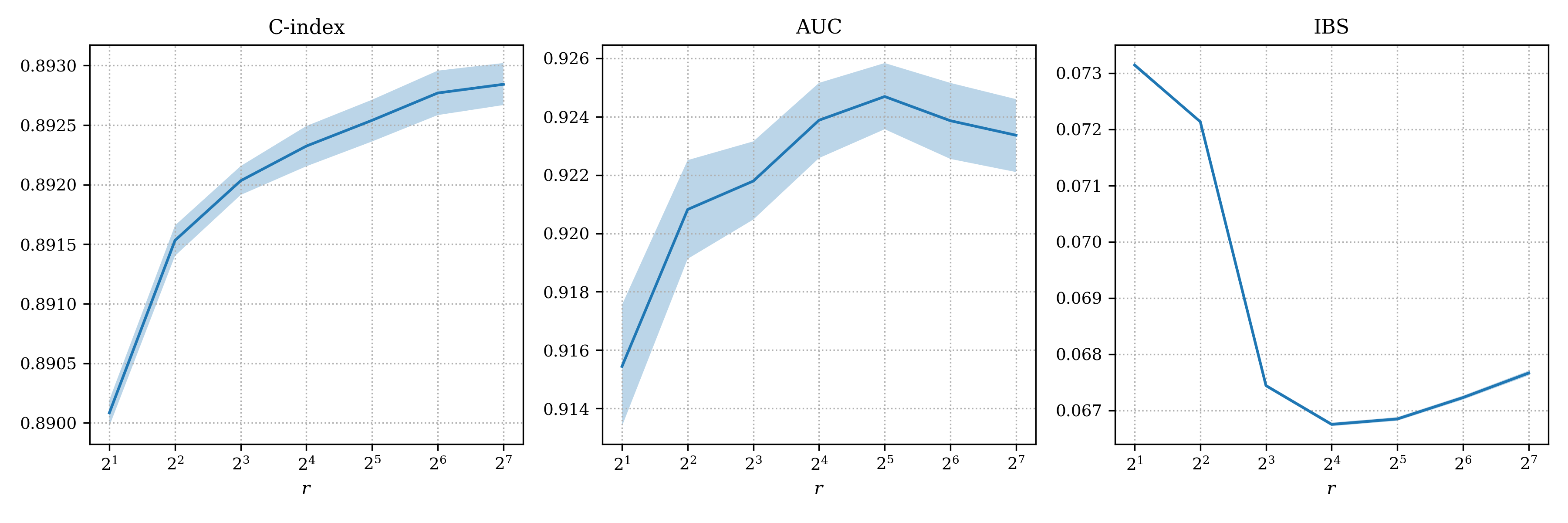}
    \caption{Influence of the number of reverse diffusion steps $r$ on C-index, integrated time-dependent AUC, and IBS. Shaded regions correspond to 95\% confidence intervals estimated from 100 repetitions.}
    \label{fig:ablation_r}
\end{figure}

Figure~\ref{fig:ablation_r} presents the mean metric values together with the corresponding 95\% bootstrap confidence intervals. The results show that the number of reverse diffusion steps affects different quality criteria in different ways. The C-index gradually increases as $r$ grows, indicating that a finer reverse-process discretization improves the ranking quality of the generated survival predictions. However, the magnitude of this improvement is relatively small, and the C-index remains within a narrow range over all tested values of $r$.

The integrated time-dependent AUC exhibits a similar but non-monotonic pattern. It increases substantially when moving from very small values of $r$ to moderate ones, reaches its maximum around $r=2^5$, and then slightly decreases for larger values of $r$. This suggests that very coarse reverse denoising is insufficient for high-quality time-dependent discrimination, while using substantially more inference steps than in the original training configuration does not provide additional benefits.

The IBS shows the clearest dependence on the reverse-process discretization. When only a few reverse steps are used, the survival probability estimates are poorly calibrated, which leads to noticeably larger IBS values. The best IBS is achieved around $r=2^4$, which is close to the original training setting of $r=20$. For larger values of $r$, the IBS slightly increases, indicating that excessive refinement of the reverse process does not necessarily improve the calibration of the estimated survival function.

Overall, these results indicate that the proposed model is reasonably robust to moderate changes in the number of reverse diffusion steps, but the best trade-off is achieved when the inference discretization remains close to the one used during training. In particular, the strongest results for AUC and IBS are obtained in the vicinity of the original training value $r=20$, while using very small values of $r$ leads to a clear degradation of survival-function calibration.

\subsubsection{Qualitative validation on synthetic survival data}

To additionally verify that the proposed model is capable of recovering the shape of an underlying continuous survival distribution, we conducted a qualitative experiment on partially synthetic data.

As the covariate space, we used the original feature vectors from the WHAS500 dataset. The observed event times were replaced with artificially generated survival outcomes. Event times were sampled according to the Cox-Weibull simulation model proposed in \cite{bender2005generating}, with regression coefficients sampled as $b \sim \mathrm{U}[-1,1]$, Weibull scale parameter $\lambda = 10^{-10}$, and shape parameter $\nu = 4$. Censoring times were generated independently from the uniform distribution on $[0, C_{\max}]$, where the value of $C_{\max}$ was selected to obtain the desired event rate.

Three censoring regimes were considered, corresponding to event rates of 25\%, 50\%, and 75\%. The resulting dataset was split into training, validation, and test subsets, after which SDPM and Random Survival Forest were trained in the same way as in the main numerical experiments.

For a randomly selected test object, the analytical survival function induced by the known Cox-Weibull model was computed and compared with the estimated survival functions obtained from RSF and SDPM. For SDPM, two representative values of the number of generated samples were considered in the visual comparison, namely $K=10^2$ and $K=10^3$. All survival curves were evaluated on a uniform time grid over the interval $[0,500]$ consisting of 1000 points.

\begin{figure}[H]
    \centering
    \includegraphics[width=0.9\linewidth]{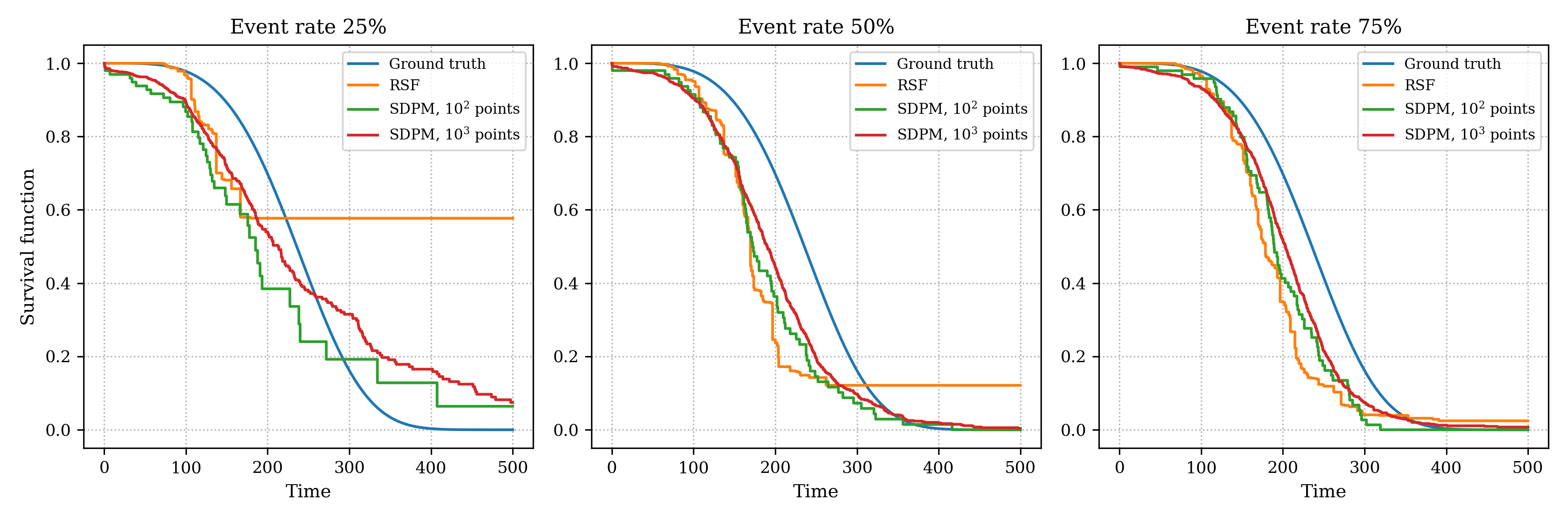}
    \caption{Qualitative comparison of survival function estimates on synthetic Cox-Weibull data for different event rates. The analytical survival function is shown together with the estimates obtained by Random Survival Forest and SDPM with $K=10^2$ and $K=10^3$ generated samples.}
    \label{fig:synthetic_sf}
\end{figure}

Figure~\ref{fig:synthetic_sf} presents the resulting survival function estimates. It can be seen that SDPM produces curves that are consistently closer to the analytical survival function than the Random Survival Forest estimate across all considered censoring regimes. RSF exhibits larger structural deviations from the reference distribution: its estimates are strongly stepwise, contain pronounced plateaus, and, depending on the censoring regime, may either underestimate survival probabilities at intermediate times or overestimate the tail of the survival curve. This behavior is especially visible in the 25\% event-rate setting, where the RSF estimate remains nearly constant at later times and does not follow the decay of the ground-truth survival function, while the SDPM curves continue to reproduce this decay more accurately.

A structural difference between the two approaches can also be observed. Since RSF constructs survival estimates as averages of Kaplan-Meier estimators, the resulting curves have jumps only at event times observed in the training data. In contrast, SDPM generates continuous-valued event times, and the corresponding Kaplan-Meier estimate is built from these generated samples. As a result, the jumps of the SDPM-based estimate are not restricted to the original training event times and can occur throughout the time axis, effectively providing a data-driven interpolation of the underlying survival distribution between observed events.

An additional effect can be observed when increasing the number of generated samples in SDPM. The estimate obtained with $K=10^3$ is smoother and visually more consistent with the analytical curve than the estimate based on $K=10^2$. This confirms that increasing the number of generated realizations improves the fidelity of the Kaplan-Meier reconstruction obtained from generated samples. This behavior highlights the ability of the proposed approach to approximate the underlying continuous survival distribution beyond the discrete event-time structure induced by the training data.

To complement the visual comparison, Table~\ref{tab:synthetic_ks} reports the average Kolmogorov-Smirnov distances between the analytical survival function and the estimated survival curves, computed over all objects in the test subset. For two survival curves $S_1(t)$ and $S_2(t)$, the reported value corresponds to
\begin{equation}
D_{\mathrm{KS}} = \max_{t \in \mathcal{G}} |S_1(t) - S_2(t)|,
\end{equation}
where $\mathcal{G}$ is the common uniform evaluation grid on $[0,500]$ consisting of 1000 points. Lower values therefore indicate a closer approximation to the ground-truth survival law.

\begin{table}[H]
    \caption{Average Kolmogorov-Smirnov distances between the analytical survival function and the estimated survival curves on synthetic Cox-Weibull data. Columns correspond to different event rates. Lower values indicate a closer approximation to the ground-truth survival distribution.}
    \centering
    \begin{tabular}{lccc}
\toprule
Model & 25\% & 50\% & 75\% \\
\midrule
RSF & 0.4424 & 0.3367 & 0.3008 \\
SDPM, $K=100$ & 0.4448 & 0.3539 & 0.2056 \\
SDPM, $K=500$ & 0.3456 & 0.3011 & 0.1803 \\
SDPM, $K=2000$ & \textbf{0.3104} & \textbf{0.2854} & \textbf{0.1683} \\
\bottomrule
\end{tabular}

    \label{tab:synthetic_ks}
\end{table}

The numerical results are consistent with the qualitative observations from Figure~\ref{fig:synthetic_sf}. For small generated sample sizes, SDPM may still be affected by sampling variability, which explains why the estimate with $K=100$ is not always better than RSF. However, as the number of generated samples increases, the Kolmogorov-Smirnov distance decreases monotonically across all considered event rates. With $K=2000$, SDPM achieves the smallest distance in all three censoring regimes, indicating a systematically better recovery of the analytical survival distribution than RSF. These results quantitatively confirm that larger generated sample sets improve the fidelity of the resulting Kaplan-Meier reconstruction.

\subsubsection{Ablation study on target-space transformations}

To assess the contribution of the proposed target-space transformations, we compared the full SDPM model with an ablated variant denoted as SDPM$_\dagger$. The ablated model is trained directly on the original event times and represents the censoring indicator using a discrete $\{-1, 1\}$ encoding. In contrast, the full SDPM model uses the transformed target representation described above, namely standardized log-transformed event times and a Gaussian-mixture representation of the censoring indicator.

The comparison is designed to evaluate not only predictive performance, but also whether the model learns a valid joint distribution of event times and censoring indicators. For this purpose, we consider three complementary criteria. First, we compare the C-index, which measures the quality of ranking induced by the generated survival estimates. Second, we evaluate event-rate calibration by comparing the fraction of uncensored generated samples with the observed event rate in the dataset. If the generated event rate is close to the observed one, this indicates that the model captures the joint structure of $(T,\delta)$ more accurately. Third, we measure the frequency of invalid or implausible generated times. Specifically, we consider negative-time outliers with $t<0$ and range-exceeding outliers with $t > 2t_{\max}$, where $t_{\max}$ denotes the maximum observed time in the dataset.

\begin{table}[H]
\caption{C-index comparison. Relative improvement is computed as $(\mathrm{SDPM} - \mathrm{SDPM}_\dagger) / \mathrm{SDPM}_\dagger$.}
\centering
\begin{tabular}{lccc}
\toprule
Dataset & SDPM & SDPM$_\dagger$ & Relative improvement \\
\midrule
FLC & 0.936 & 0.936 & +0.1\% \\
Ovarian & 0.635 & \textbf{0.640} & -0.8\% \\
PBC & \textbf{0.997} & 0.996 & +0.1\% \\
Retinopathy & \textbf{0.571} & 0.565 & +1.1\% \\
Rotterdam & \textbf{0.724} & 0.710 & +2.1\% \\
SEER & \textbf{0.753} & 0.752 & +0.1\% \\
SUPPORT & \textbf{0.894} & 0.860 & +3.9\% \\
TCGA-GBM & \textbf{0.891} & 0.885 & +0.6\% \\
VLBW & \textbf{0.878} & 0.877 & +0.2\% \\
WHAS500 & 0.755 & \textbf{0.760} & -0.7\% \\
\midrule
Average & -- & -- & \textbf{+0.7\%} \\
\bottomrule
\end{tabular}
\label{tab:c-index}
\end{table}

The effect of the transformations on ranking quality is reported in Table~\ref{tab:c-index}. SDPM improves the C-index on eight out of ten datasets, with an average relative improvement of 0.7\%. The largest gain is observed on SUPPORT, where the relative improvement reaches 3.9\%. Although the improvement is moderate in magnitude, it is consistent across most datasets, indicating that the transformed target representation does not merely improve sample validity, but also benefits predictive discrimination.

\begin{table}[H]
\caption{Event-rate calibration across datasets. Values are percentages; bold indicates the model closer to the observed event rate. Average error is the mean absolute error in percentage points.}
\centering
\begin{tabular}{lccc}
\toprule
Dataset & Observed & SDPM & SDPM$_\dagger$ \\
\midrule
FLC & 27.6\% & 27.3\% & \textbf{27.5\%} \\
Ovarian & 59.6\% & \textbf{87.0\%} & 88.2\% \\
PBC & 38.5\% & 41.3\% & \textbf{40.0\%} \\
Retinopathy & 39.4\% & \textbf{44.2\%} & 47.9\% \\
Rotterdam & 42.6\% & 47.0\% & \textbf{41.5\%} \\
SEER & 15.3\% & \textbf{15.2\%} & 15.8\% \\
SUPPORT & 68.1\% & \textbf{69.9\%} & 75.1\% \\
TCGA-GBM & 82.4\% & 84.6\% & \textbf{81.6\%} \\
VLBW & 17.3\% & \textbf{14.7\%} & 12.8\% \\
WHAS500 & 43.1\% & \textbf{46.6\%} & 57.5\% \\
\midrule
Average error & -- & \textbf{5.0\%} & 6.7\% \\
\bottomrule
\end{tabular}
\label{tab:event-rate}
\end{table}

Table~\ref{tab:event-rate} shows that the proposed target-space transformations improve the calibration of the generated censoring structure. SDPM is closer to the observed event rate on six out of ten datasets and achieves a lower average absolute error than SDPM$_\dagger$: 5.0 percentage points compared with 6.7 percentage points. This suggests that the transformed representation helps the model better approximate the joint distribution of event times and censoring indicators, rather than only producing accurate rankings.

\begin{table}[H]
\caption{Outlier rates across datasets. Values are percentages; bold indicates the lower outlier rate. Negative-time outliers correspond to $t < 0$, while range-exceeding outliers correspond to $t > 2t_{\max}$, where $t_{\max}$ is the maximum observed time in the dataset.}
\centering
\begin{tabular}{lcccc}
\toprule
\multirow{2}{*}{Dataset}
& \multicolumn{2}{c}{Negative-time}
& \multicolumn{2}{c}{Range-exceeding} \\
\cmidrule(lr){2-3}
\cmidrule(lr){4-5}
& SDPM & SDPM$_\dagger$ & SDPM & SDPM$_\dagger$ \\
\midrule
FLC & \textbf{0.0\%} & 0.1\% & 0.0\% & 0.0\% \\
Ovarian & \textbf{0.0\%} & 7.4\% & 0.1\% & \textbf{0.0\%} \\
PBC & \textbf{0.0\%} & 1.8\% & \textbf{0.1\%} & 0.7\% \\
Retinopathy & \textbf{0.0\%} & 25.9\% & \textbf{13.0\%} & 16.0\% \\
Rotterdam & \textbf{0.0\%} & 7.0\% & 0.0\% & 0.0\% \\
SEER & 0.0\% & 0.0\% & 0.0\% & 0.0\% \\
SUPPORT & \textbf{0.0\%} & 28.2\% & 0.0\% & 0.0\% \\
TCGA-GBM & \textbf{0.0\%} & 2.3\% & \textbf{0.2\%} & 1.0\% \\
VLBW & \textbf{0.0\%} & 24.0\% & 0.0\% & 0.0\% \\
WHAS500 & \textbf{0.0\%} & 15.8\% & \textbf{0.4\%} & 0.7\% \\
\midrule
Average & \textbf{0.0\%} & 11.3\% & \textbf{1.4\%} & 1.8\% \\
\bottomrule
\end{tabular}
\label{tab:time-outliers}
\end{table}

Table~\ref{tab:time-outliers} further demonstrates that the proposed transformations substantially improve the validity of generated event times. The full SDPM model eliminates negative-time outliers on all datasets, whereas SDPM$_\dagger$ produces such invalid samples on most datasets, with an average rate of 11.3\%. This confirms the importance of modeling event times in the transformed space, where the support constraints are easier to satisfy. For range-exceeding outliers, both models produce relatively low rates on average; however, SDPM still achieves a lower mean outlier rate, 1.4\% compared with 1.8\% for SDPM$_\dagger$.

Overall, the ablation study confirms that the proposed target-space transformations are important for stable and meaningful generative modeling of survival outcomes. They improve the calibration of the generated censoring structure, reduce the number of invalid generated times, and provide a consistent improvement in C-index on most datasets. These results support the use of standardized log-transformed times and a continuous mixture-based representation of the censoring indicator in the final SDPM model.

\section{Conclusion}

In this paper, we proposed the Survival Diffusion Probabilistic Model (SDPM), a generative approach to continuous-time survival analysis. Instead of directly parameterizing the hazard or survival function, SDPM models the conditional distribution of the observable survival outcome, represented by the observed time and censoring indicator, $\mathbb{P}(T,\delta \mid \mathbf{x})$. The learned distribution is used to generate conditional samples of observed survival outcomes, which are then converted into individual survival function estimates using the Kaplan-Meier estimator.

The proposed formulation combines the flexibility of diffusion-based generative modeling with the robustness of classical nonparametric survival estimation. SDPM operates in continuous time and avoids a fixed discretization of the output time axis. To improve stability and validity of generated outcomes, the model uses standardized log-transformed times and a continuous Gaussian-mixture representation of the censoring indicator.

Experiments on ten real-world survival datasets demonstrate that SDPM is competitive with strong tree-based, boosting-based, and neural survival baselines. The method achieves strong results in terms of Harrell's C-index and integrated time-dependent AUC, and shows its clearest advantage in terms of the integrated Brier score. This indicates that SDPM is particularly effective at estimating calibrated survival functions, not only at ranking individuals by risk.

The additional experimental analysis supports the main motivation of the method. Increasing the number of generated samples improves the stability and quality of the Kaplan-Meier reconstruction, although it also increases inference time. The study of the number of reverse diffusion steps shows that the model is reasonably robust to moderate changes in the sampling discretization, with the best trade-off obtained near the training configuration. Experiments on synthetic Cox-Weibull data further demonstrate that SDPM can recover the shape of an underlying continuous survival distribution and that larger generated sample sets lead to a closer approximation of the analytical survival law.

The ablation study confirms that the proposed target-space transformations are important for stable generative modeling of survival outcomes. Compared with a variant trained directly on the original time scale and a discrete censoring encoding, the full SDPM model improves event-rate calibration, eliminates negative generated times, reduces range-exceeding outliers, and provides consistent gains in C-index on most datasets.

A limitation of SDPM is its sampling-based inference procedure: estimating a survival function requires generating multiple conditional realizations and applying the Kaplan-Meier estimator, which introduces additional computational cost. However, the experiments show that this cost can be controlled by selecting a moderate number of generated samples while preserving strong predictive performance.

Future work includes developing faster sampling procedures for survival diffusion models, investigating continuous-time score-based formulations, and extending the proposed approach to more complex survival settings, including competing risks, recurrent events, and multivariate event outcomes.

\newpage
\bibliographystyle{unsrt}
\bibliography{TabTrans,Survival_analysis,Surv_Attent,Stas}

\newpage
\appendix
\section{Hyperparameter search spaces}\label{sec:hyperparams}

This appendix summarizes the hyperparameter search spaces used in Optuna for all compared models. For all methods, hyperparameter optimization was performed independently within each training fold. The search spaces were defined manually for each model according to its architecture and training procedure.

\subsection{DeepSurv}

For DeepSurv, the following hyperparameters were optimized:
\begin{itemize}
    \item number of hidden layers: $\{1,2,3,4\}$;
    \item hidden layer dimension: $\{64,128,256,512\}$;
    \item activation function: \{ReLU, SiLU, Tanh\};
    \item learning rate: log-uniform on $[10^{-4}, 5 \cdot 10^{-3}]$;
    \item batch size: $\{32,64,128,256\}$;
    \item normalization: \{False, True\};
    \item dropout mode: \{zero, nonzero\};
    \item dropout rate: log-uniform on $[10^{-2}, 2 \cdot 10^{-1}]$ if nonzero dropout was selected, and $0$ otherwise.
\end{itemize}

\subsection{DeepHit}

DeepHit used the same search space as DeepSurv, with two additional hyperparameters:
\begin{itemize}
    \item $\alpha$: uniform on $[0,1]$;
    \item $\sigma$: $\{10^{-2}, 10^{-1}, 0.5, 1.0, 2.0\}$.
\end{itemize}

\subsection{SDPM}

For the SDPM model, the following hyperparameters were optimized:
\begin{itemize}
    \item batch size: $\{32,64,128\}$;
    \item learning rate: log-uniform on $[10^{-4}, 5 \cdot 10^{-3}]$;
    \item weight decay: $\{0, 10^{-6}, 10^{-5}, 10^{-4}\}$;
    \item dropout rate: $\{0\}\cup \mathrm{U}[0.05,0.25]$;
    \item number of hidden layers: $\{2,3,4,5\}$;
    \item hidden dimension: $\{64,128,256,512\}$;
    \item number of Fourier frequencies $m$: $\{8, 16, 32, 64\}$;
    \item Fourier frequency initialization scale: log-uniform on $[0.01, 1]$;
    \item activation function: \{ReLU, SiLU\};
    \item normalization type: \{do not use layer normalization, layer, conditional (AdaLN-Zero)\};
    \item categorical feature processing: \{embedding, empty (without processing)\};
    \item categorical embedding dimension: $\{4,8\}$ when embeddings were used;
    \item number of reverse diffusion steps $r$: $\{10,20,30\}$;
    \item diffusion time embedding dimension: $\{8,16\}$;
    \item cosine scheduler parameter $s$: log-uniform on $[10^{-4}, 1.5\cdot10^{-1}]$;
    \item noise embedding flag: \{False, True\}.
\end{itemize}

Total number of epochs was fixed to 1000 for DeepSurv, DeepHit, and SDPM.

\subsection{Random Survival Forest}

For Random Survival Forest, the following hyperparameters were optimized:
\begin{itemize}
    \item number of trees: $\{50,100,200,400,800\}$;
    \item maximum depth: \{$\infty$, 4, 8, 12, 16\};
    \item minimum samples to split: $\{2,4,8,16,0.01,0.02,0.05\}$;
    \item minimum samples per leaf: $\{1,2,4,8,0.005,0.01,0.02\}$;
    \item maximum number of features: \{sqrt, log2\};
    \item maximum number of leaf nodes: \{$\infty$, 32, 64, 128, 256\};
    \item bootstrap: \{True, False\}.
\end{itemize}

\subsection{XGBSE-based models}

Both XGBSE-based models shared a common gradient boosting search space:
\begin{itemize}
    \item early stopping rounds: $\{10,25,50\}$;
    \item learning rate: log-uniform on $[10^{-3}, 2 \cdot 10^{-1}]$;
    \item maximum depth: $\{2,3,4,5,6,8\}$;
    \item minimum child weight: log-uniform on $[10^{-1}, 20]$;
    \item subsample ratio: uniform on $[0.5,1.0]$;
    \item column subsample ratio: uniform on $[0.5,1.0]$;
    \item $\ell_2$ regularization: log-uniform on $[10^{-4}, 10]$;
    \item $\ell_1$ regularization: log-uniform on $[10^{-4}, 10]$;
    \item AFT loss distribution: \{normal, logistic, extreme\};
    \item AFT loss distribution scale: log-uniform on $[0.1, 10]$.
\end{itemize}

For XGBSEStackedWeibull, the following additional hyperparameters were optimized:
\begin{itemize}
    \item Weibull penalty: log-uniform on $[10^{-2}, 10]$;
    \item Weibull $\ell_1$ ratio: uniform on $[0,1]$.
\end{itemize}

For XGBSEKaplanNeighbors, the following additional hyperparameter was optimized:
\begin{itemize}
    \item number of neighbors: $\{10,20,\dots,100\}$.
\end{itemize}

For both XGBSE-based models, the number of boosting rounds was fixed to 1000.
\end{document}